# Investigating Critical Risk Factors in Liver Cancer Prediction

Jinpeng Li, Yaling Tao, and Ting Cai

**Abstract**—We exploit liver cancer prediction model using machine learning algorithms based on epidemiological data of over 55 thousand peoples from 2014 to the present. The best performance is an AUC of 0.71. We analyzed model parameters to investigate critical risk factors that contribute the most to prediction.

✦

## 1 INTRODUCTION

LIVER cancer is a threat to human health and life over the world. According to recent statistics, the crude incidence of liver cancer ranks top five among all cancers in China, and the death rate ranks the top two [1]. A large number of studies have shown that the occurrence of liver cancer is closely related to certain risk factors such as living habits and disease history [2], [3], [4]. If people could pay enough attention to risk factors and adjust their lifestyles accordingly, liver cancer can be prevented to a large extent [5]. Therefore, investigating critical risk factors and quantitative assessment of risk factors are beneficial to personal cancer prevention and public health management so as to promote human well-being.

In 2012, China National Cancer Center and Cancer Hospital of Chinese Academy of Medical Sciences carried out Early Diagnosis and Treatment of Urban Cancer (EDTUC) project, and released risk factors for the top-five common cancers in China, where liver cancer was one of them. Risk factors were identified to be suitable to Chinese people by experienced epidemiologists and clinical oncologists with reference to Harvard Cancer Risk Index (HCRI) [6]. EDTUC risk factors were contained in a questionnaire to be collected in major cities, where Ningbo is one of them. To ensure sufficiency and manageability, the questionnaire involved a broad range of non-clinical information (around one hundred risk factors), which is more than existing researches [2], [3], [7]. Based on epidemiological knowledge, each questionnaire was tagged with a suspected label on whether a person has high probability of developing liver cancer. From 2014 to the latest present, the clinical diagnosis records have been accumulated gradually, which could be regarded as diagnosis label concerning whether liver cancer eventually occurred within five years on each questionnaire-participant. Based on diagnostic labels, it is now possible to analyze the relationship between each factor and diagnostic results in a goal-driven approach.

We retrieve diagnostic records from Ningbo Health Commission (NHC), China and exploit machine learning models to predict liver cancer based on EDTUC risk factors. Different from classical statistical analysis [8], [9], we identify critical risk factors by assessing the contribution score of each risk factor in decision-making. This goal-driven method provides supplemental information to statistical analysis while achieving certain goals.

As Fig. 1 shows, EDTUC questionnaire considers six aspects of information, and each aspect includes multiple risk factors. We concatenate information of all risk factors as a high-dimensional risk vector, which is the input to Liver Cancer Prediction Model (LCPM). LCPM is implemented with various classical machine learning models and Deep Neural Network (DNN). After supervised training, LCPM learns decision rules to predict liver cancer. Afterwards, we analyze LCPM parameters to rank the contribution scores of risk factors, where critical risk factors are with leading scores.

This paper initializes the research on liver cancer prediction and risk factor identification in EDTUC project based on up-to-date diagnostic information in Chinese people. According to our results, machine learning models show potential in successful prediction of liver cancer within five years. Meanwhile, most identified critical risk factors have been intensively studied in the past, whereas some have not. We suggest more medical and epidemiological researches on these factors.

## 2 RELATED WORKS

According to Global Cancer Statistics (2012), liver cancer among males is a leading cause of cancer death in developing countries and regions [1]. Liver cancer is predominant in Asian countries including China [7]. It has been demonstrated that liver cancer could be prevented by applying prevention measures, such as tobacco control, alcohol control, and the usage of early detection tests [1]. Identifying more risk factors leading to liver cancer occurrence is an important research topic.

_________________
*J. Li, Y. Tao, and T. Cai are with Ningbo HwaMei Hospital, University of Chinese Academy of Sciences, Ningbo, Zhejiang province, P. R. China (e-mail: {lijinpeng, taoyaling, caiting}@ucas.ac.cn).*

*J. Li and T. Cai are also with Institute of Life and Health Industry, University of Chinese Academy of Sciences, No. 41 Northwest Street, Ningbo 315010, Zhejiang Province, P. R. China.*

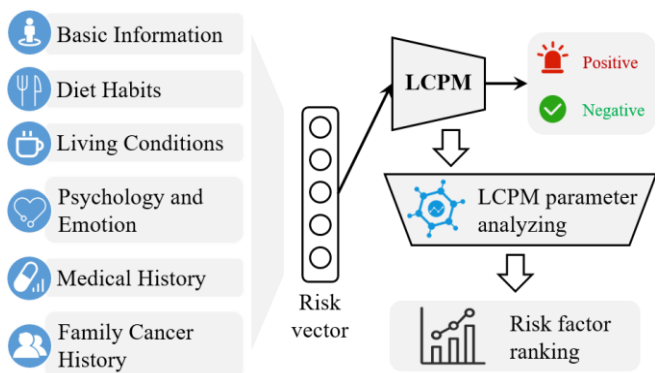

Fig. 1. LCPM working scenario and critical risk factor identification. LCPM considers six aspects of information and makes predictions on five-year liver cancer occurrence. Critical risk factors are identified as high-ranking factors ranked by analyzing model parameters.

Liver cancer includes many species, where hepatocellular carcinoma (HCC) is the predominant type in China and other Asian countries [10]. Until now, most existing researches have focused on HCC to find critical risk factors of liver cancer. We summarize and divide risk factors that have been proved to be associated with liver cancer (or specifically HCC if specified) into four categories.

The first is basic risk factor. Gender is associated with HCC, where males are 2.4 times more likely than females to develop HCC in the worldwide distribution [7]. Age is a critical factor in the development of liver cancer. According to [11], the HCC incidence increases with age, reaching highest prevalence among those aged over 65 years. Obesity have been recognized as a significant risk factor for liver cancer, where the relative risks (RR) for liver cancer for the overweight (BMI $\geq$ 30) were 1.89 (95% CI = 1.51-2.36) [3].

The second is viral infection. Hepatitis-B Virus (HBV) has been substantially proved to be the most common cause of HCC worldwide, which accounts for over 50% of all liver cancers via statistical analysis [7], [12]. Hepatitis-C Virus (HCV) ranks the second in all the causes of HCC. According to [13], chronic HCV infection is associated with a 20-30-fold increased risk of developing HCC as compared to uninfected individuals.

The third is personal living habits. Strong evidence has been found between alcohol drinking and HCC. For people drinking more than 60 g alcohol per day, the effect was more pronounced [7]. Heavy alcohol intake (210-250 g per week) increase HBV patients' HCC risk [14], [15]. Existing researches have also established association between tobacco smoking and HCC occurrence [15]. In addition, the lack of physical exercise is also an important factor in the occurrence of liver cancer [16].

The forth is hepatobiliary system diseases. Alcohol-related cirrhosis has been proved to be a major cause of HCC in populations with low prevalence of HBV and HCV infection [3]. Fatty liver and or non-alcohol fatty liver disease (non-alcoholic steatohepatitis) are tightly related to HCC [17].

Most existing works were based on statistical methods to find the correlations between variables [3], [8]. Machine learning is another approach of data miming, which models the functional relationships between variables while completing certain tasks. The decision-making process of machine learning models sheds light on the relative contributions of risk factors rather than considering them independently.

Deep learning (DL) [18] is a class of machine learning model characterized with learning hierarchical representations optimized with goal-driven training, and DNN is one type of DL. DNN has been widely used in the medical field and demonstrated its advantage in medical image analysis [19] and electronic health records analysis [20], etc. These scenarios occur after the patient enters the hospital, whereas disease cannot be predicted in advance. Concerning DL-based cancer prediction, most works have been focused on analyzing genetic information [21], [22] or medical images [23], [24], which require substantial clinical measures. Few works have been conducted to predict cancers using epidemiological risk factors alone [25]. To our knowledge, there is a lack of in-depth study to predict liver cancer (or HCC) based on epidemiological factors. Several works have used saliency maps to detect key information in decision-making [26], however, they were designed for medical images. Saliency maps changes with the input so as not to measure feature importance on fixed input dimension indexes.

In this paper, since the input is high-dimensional risk vector, we use DNN to implement LCPM, which show superiority over other considered machine learning models. We then analyze the model parameters to identify critical risk factors contributing the most in prediction.

## 3 MATERIALS AND METHODS

We retrieve and preprocess EDTUC data and corresponding clinical diagnostic information, which are used to train LCPM via supervised training. We analyze LCPM parameters to locate critical risk factors, which are then used to train new LCPM.

### 3.1 Data Retrieval and Preprocessing

EDTUC project uses epidemiological questionnaire to assess common cancer risks. The participants satisfy the following conditions simultaneously: The age is between 40 and 74 (both male and female); Voluntary and able to accept the questionnaire; No serious organ dysfunction or mental diseases; No history of tumor diagnosis and no serious endoscopic disease being treated. EDTUC have been carried out in major cities of China including Ningbo. In order to promote the project among large groups of people, EDTUC questionnaires are mainly collected at community hospitals and regional central hospitals. Each questionnaire is recorded by a dedicated person using Excel form, which is then examined by an independent checker. The Ethical Approval Number is 15-070/997 granted by Ethics Committee of Chinese Academy of Medical Sciences. All the questionnaires across Ningbo are gathered at NHC.

In the original version of EDTUC questionnaire, there were more than one hundred questions, which involves logical repetitions. To compress risk factors into compact vectors while maintaining information (beneficial to the training process of machine learning models), we use 84 risk factors to predict liver cancer. Due to space limitations, we post these risk factors on the Github website (https://github.com/JPLi1109/Liver-Cancer-Prediction). We retrieve samples between 2014 and 2017 from NHC and obtain 55891 unique questionnaire samples.

The questionnaire has choice questions and fill-in questions. The choice questions has two types: single-choice questions and multiple-choice questions. For each single-choice question, we use a single scalar started from zero to represent choice. For each multiple-choice question, we use a high-dimensional vector to represent choices, where each dimension corresponds to a choice (either zero or one). For fill-in questions, we use the Arabic numbers directly to represent content. As Fig. 1 shows, we concatenate the results of all questions to produce the input risk vector. We denote risk vector as $x \in R^N$, where $N$ is the vector dimension, i.e., number of risk factor. In order to avoid dimensions with larger scales dominate the optimization, each dimension is scaled to [0, 1] according to

$$x_i = \frac{x_i^{max} - x_i}{x_i^{max} - x_i^{min}}, \quad (1)$$

where $x_i$ is the $i^{th}$ risk factor; $x_i^{max}$ and $x_i^{min}$ are maximum and minimum values of $i^{th}$ risk factor in dataset.

We use name and identity number to search for diagnostic information at NHC database from 2014 to the present. According to the International Classification of Diseases 10th revision (ICD-10), the general code of liver cancer is C22, C23, and C24 and HCC is the most common one. We study liver cancer in a broad sense, and retrieved all the results with ICD-10 code C22, C23, and C24. There are 184 participants diagnosed with liver cancer from January 2014 to September 2019. These samples are tagged with positive label, and the other samples are tagged with negative label. In terms of the longest time span, this study involves prediction of liver cancer within five to six years.

### 3.2 Model Training

LCPM is a binary classifier determining whether a risk vector is at high probability of developing liver cancer. We compare the performance of various machine learning methods in terms of True Positive Rate (TPR, sensitivity), False Positive Rate (FPR), True Negative Rate (TNR, Specificity), False Negative Rate (FNR), Precision, F1-score, and Accuracy. We use several models to implement LCPM: Logistic Regression (LR). LR has been widely used in both statistical analysis and machine learning to model the relationship of variables. Existing works have applied LR to predict diseases [27]. LR computes the positive probability while making decisions via Sigmoid function. Decision Tree (DT). DT evaluates information gain of each feature dimension, and uses tree-like model for decision consequences [28]. Support Vector Machine (SVM). Unlike LR, it considers samples close to the decision boundary rather than considering all samples. With the cooperation of kernel functions, SVM is able to deal with nonlinear classifications. AdaBoost [29] dynamically adjusts the weights of samples and weak classifiers to achieve strong classifiers, and we introduce it as a comparing method. Different from the above models, DNN learns efficient representations of input in a hierarchical structure and thus allows low-level representations to evolve into high-level representations along multiple layers. DNN is typically optimized using goal-driven training. We use Sigmoid function to normalize outputs, which enables DNN to compute probabilities during decision-making.

### 3.3 Identifying Critical Risk Factors

We identify critical risk factors in the risk vector by analyzing the weight distributions of DNN. During the training process, the weights are optimized to minimize classification error. Risk factors with relatively high contribution in liver cancer prediction tend to evolve higher weights. Since DNN has many layers, hidden layers are not directly linked with input dimensions. Therefore, we only analyze input-layer weights to assess the feature contribution. Considering the absolute value of weight is more important than its positive and negative attribute, we take the absolute value of the first layer weight. In addition, informative dimensions tend to have high deviation. Therefore, we use the following to compute contribution scores of each dimension:

$$C_i = \sigma_i \sum_{j=1}^{H} |w_{ij}|, \quad (2)$$

where $C_i$ denotes the contribution of the $i^{th}$ risk factor, $\sigma_i$ is the sample deviation of the $i^{th}$ risk factor, $H$ is the number of neurons of the first hidden layer, and $w_{ij}$ means the connection weight between $i^{th}$ risk factor and $j^{th}$ neuron in the first hidden layer. In LR, we use the coefficients between risk factors to evaluate relative contributions of risk factors.

Then, we use the identified critical risk factors to train new LCPM, which is compared with LCPM using all risk factors. In practice, we could use fewer risk factors to acquire LCPM, which is more convenient to users.

## 4 EXPERIMENTS AND RESULTS

### 4.1 LCPM with All Risk Factors

After data retrieval and preprocessing, 55891 samples with a dimension of 84 are obtained, where 184 samples have been tagged with positive label and the rest 55707 samples are labeled as negative. We first divide data as unbalanced training set (80%) and unbalanced test set (20%) randomly. The severe sample imbalance between positive and negative samples in the training set would mislead LCPM to yield negative predictions. Therefore, we apply random undersampling technique to re-balance the training set. We repeat the experiments ten times, and analyze averaged results on the unbalanced test set. In each experiment, we randomly select 184 negative samples (equal to the positive samples) to train the model. Before model training, we normalize the training set and test set to [0,1], independently.

TABLE 1
PREDICTING LIVER CANCER USING DIFFERENT MACHINE LEARNING METHODS

| Method | Sensitivity | FPR | Specificity | FNR | Accuracy | AUC |
|---|---|---|---|---|---|---|
| Baseline | 30.56 | 11.22 | 88.78 | 69.44 | 88.59 | None |
| Logistic Regression | 69.51 (4.67) | 37.46 (0.56) | 62.55 (0.56) | 30.49 (4.67) | 62.57 (0.57) | 0.718 (0.021) |
| KNN | 61.59 (6.71) | 44.02 (2.10) | 55.98 (2.10) | 38.41 (6.72) | 56.00 (2.09) | None |
| Support Vector Machines | 73.78 (5.41) | 42.10 (2.67) | 57.90 (2.67) | 26.22 (5.41) | 57.98 (2.72) | 0.698 (0.010) |
| Decision Tree | 62.81 (3.65) | 46.19 (2.76) | 53.81 (2.76) | 37.20 (3.65) | 54.32 (2.63) | None |
| AdaBoost | 59.15 (2.34) | 46.35 (3.11) | 53.65 (3.11) | 40.85 (2.34) | 53.60 (3.11) | None |
| Deep Neural Networks | 73.78 (3.66) | 39.90 (4.42) | 60.09 (4.43) | 26.22 (3.66) | 60.15 (4.41) | 0.713 (0.031) |

*Experiments have been repeated ten times. Baseline: Using suspected label to predict liver cancer. Results shown in mean (standard deviation).*

Each sample have been evaluated by experts on liver cancer risk using human experience and knowledge. Before LCPM training, we first use suspected labels (see Section. I) to predict liver cancer, which acts as baseline model to LCPM. The results appear at the first row of Table. I named 'Baseline'.

LCPM training details are as follows. In SVM, we apply RBF kernel function to encourage generalization. No class weights have been introduced. To search for hyperparameters $c$ and $gamma$, we use Grid Search and threefold cross-validation to maximize F1-Score, and the searching scope are both $[2^{-3}, 2^3]$. In DT, we use Gini Impurity as criterion and tree model is set to expand until all leaves are pure. In AdaBoost, we use SAMME algorithm with a maximal depth of 10 and the minimal sample split is set to 10, and the estimator number is 200. The above models are implemented with Scikit-learn Library, and the results are summarized in Table. I.

We use TensorFlow Library to construct DNN. After trying different layer number and neuron numbers, we come to the following light DNN: The input layer has 84 neurons, which is equal to the dimension of risk vectors. The second layer and the third layer both have eight neurons, and the output layer has one neuron. The output activation is Sigmoid and the rest layers have Rectified Linear Unit (ReLU) activations. L2-regularizer (0.02) is applied on the second and third layer to encourage the smoothness of fitting. We use Adam optimizer with default parameters ( $learning\ rate = 0.001, \beta_1 = 0.9, \beta_2 = 0.999$). No AMSGrad is applied. The batch size is set to 32. In each training epoch, we use 20% of training data as validation data. We apply early-stopping technique to training process, where patience value is set to 10.

Baseline results are obtained by comparing suspected labels and diagnostic labels. Although the FRP, Specificity and Accuracy are reasonable in Baseline model, the sensitivity, Miss Rate, Precision and F1-Score are poor. These phenomena indicate that suspected labels fail to predict liver cancer within five years. The data-driven LR, SVM, DT and AdaBoost models all yield reasonable results. Concerning sensitivity, DNN shows significant advantage over other models ($p < 0.01$ to KNN, DT and AdaBoost; $p < 0.05$ to LR). No significant differences in sensitivity have been detected between DNN and SVM ($p > 0.1$). Concerning FPR, DNN shows advantage over KNN ($p < 0.01$), DT ($p < 0.01$), AdaBoost ($p < 0.01$) and SVM ($p < 0.05$), whereas no significant advantage has been found over DNN and LR ($p > 0.1$). In addition, the specificities of DNN and LR are significantly higher than those of the other methods. The miss rates of DNN, SVM and LR are significantly lower than those of KNN, DT and AdaBoost. DNN, LR, and SVM are suitable models in our task. The mean AUC of SVM is lower than those of LR and DNN. Considering the above results, LR and DNN achieve the best results in liver cancer prediction.

### 4.2 Critical Risk Factors Identification

Results have proved that DNN and LR are reliable LCPMs, where DNN shows better sensitivity. We use (2) to compare relative contributions of risk factors in DNN, which measures the scaled and absolute weight distribution of the input layer to locate high-contributing dimensions in risk vector during decision-making. The results are shown in Fig. 2. Results obtained by analyzing LR coefficients are also provided in Fig. 2 for reference. The relative importance among risk factors shows some consistency in LR and DNN. However, DNN gives greater weight to diet habits and living conditions. In diet habits, B02: Fresh Fruit and B07: Salt Intake have high scores in DNN, whereas low in LR. In living conditions, C01: Air Pollution, C05: Smoking, C09: Regular Inhalation of Secondhand Smoke, and C13: Regular Physical Exercise have high scores in DNN, whereas the scores are low in LR. The top 10% risk factors (nine factors) are: A02: Age, A01: Gender, C01: Air Pollution, A07: Occupation, C09: Regular Inhalation of Secondhand Smoke, E32: Hyperlipidemia, C13: Regular Physical Exercise, D02: Long-Term Mental Depression, and F01: Cancer History of Blood Relatives.

### 4.3 LCPM with Critical Risk Factors

Critical risk factors are assumed to have higher contributions in DNN classification. To confirm this assumption, we rank risk factors according to contribution scores and use top 10% (nine risks), 25% (21 risks), 50% (42 risks), and 75% (63 risks) factors to train new DNN-based LCPMs. The new LCPMs have the same structure as original LCPM except for input neuron numbers. Perfor-mance of these LCPMs are compared to that of LCPM using all risk factors in terms of sensitivity and FPR (two major measures in liver cancer prediction). The results are summarized in Fig. 3. Using top 10% risk factors, the sensitivity is 68.33%, which drops by around 5.5% comparing with that using all risk factors ($p < 0.01$). Using 50% factors and 75% factors have statistically similar results to those using all risk factors. Concerning FPR, LCPM using 10% factors has even lower FPR comparing with that using all factors ($p < 0.01$). FPR also prevails when using 25% risk factors ($p < 0.01$). Using 50% risk factors, FPR shows no significant difference comparing with that using all risk factors. FPR using 75% risk factors is higher than that using all risk factors ($p < 0.01$). Therefore, using few critical risk factors (10%, 25%) reduces FPR at the cost of reducing sensitivity. Both sensitivity and FPR show reasonable results when predicting liver cancer with only few critical risk factors, which validates of our method in that they contain discriminative information.

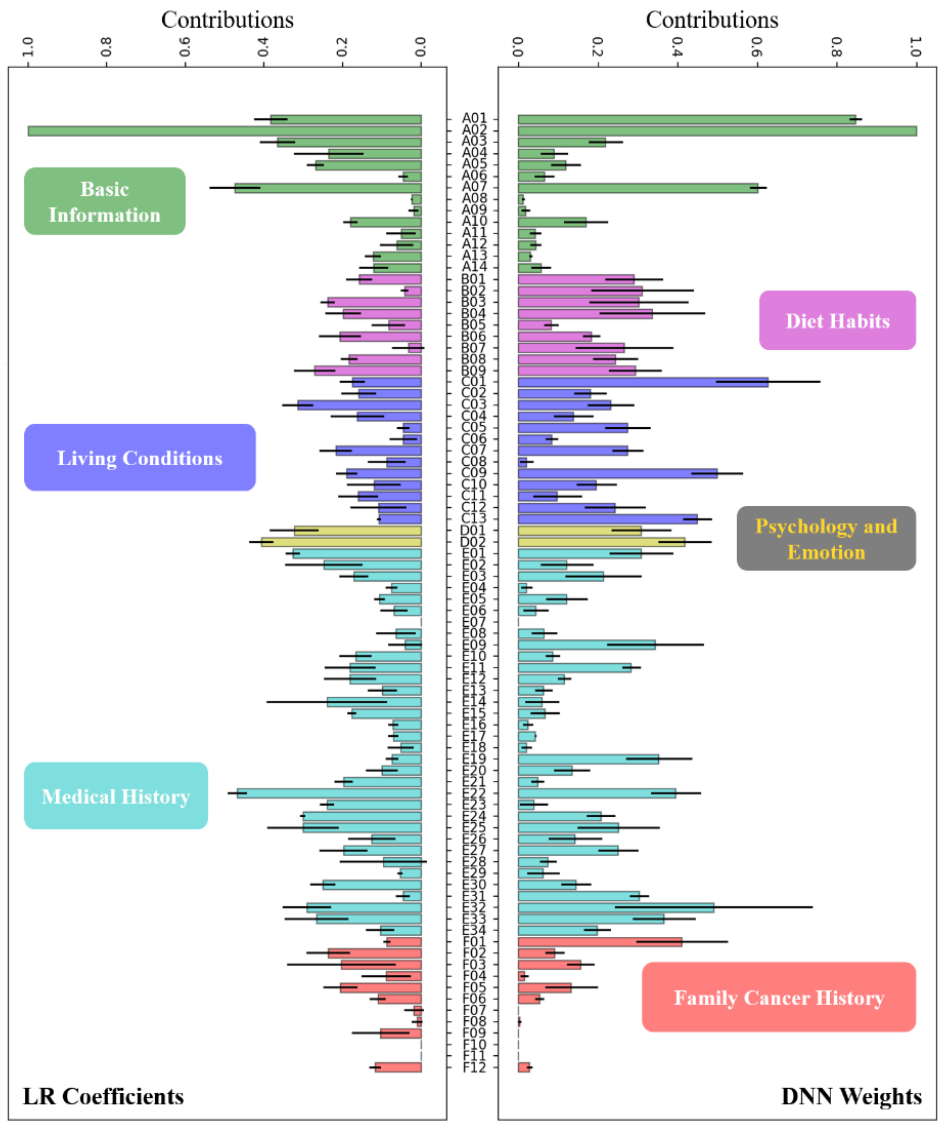

Fig. 2. Risk factor scores using LR coefficients and DNN weights. Experiments have been repeated ten times (ten LR models and DNN models obtained), and error bars show standard deviation. A: Basic Information; B: Diet Habits; C: Living Conditions; D: Psychology and Emotion; E: Medical History; F: Family Cancer History.

## 5 Discussions

The baseline has high Specificity (88.78%) and low Sensitivity (30.56%), which means only 30.56% high-risk cases are successfully identified and 88.75% low-risk cases are successfully excluded from liver cancer. In cancer prediction, identifying high-risk groups as completely as possible is more important than excluding low-risk groups as completely as possible. Therefore, the baseline model failed to predict liver cancer. The reason for this phenomenon is probably because baseline model makes decisions based on only a few high-risk factors (e.g., tobacco, age), and it is difficult to comprehensively consider the coupling of multiple factors in decision-making.

DNN achieves good performance in liver cancer prediction. To explain the inner working mechanism of the parametric model, we visualize the input risk vector, and the hidden layer representations using Principal Compo-nent Analysis (PCA) to find some clues. Fig. 4 shows the 2-D projection of input risk vectors and hidden representations generated in hidden layers using two leading components. Representations become more linearly separable in deeper layers. This phenomenon implies that when our DNN performs classification task, it increases the linear separability of representations layer by layer, which is conducive to the final pattern classification. This phenomenon may explain the advantage of DNN in our task although it appears as a black-box model.

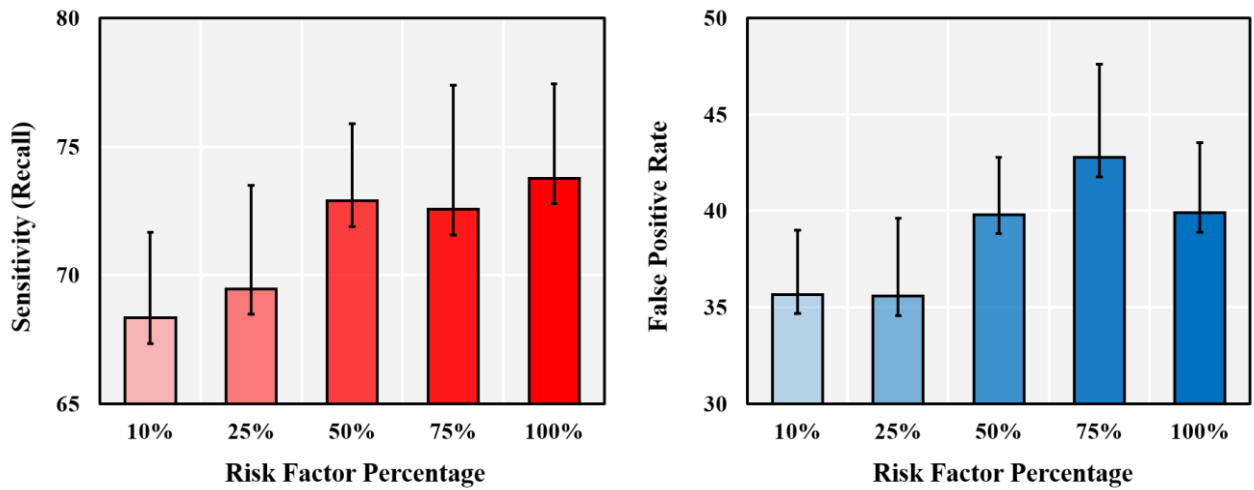

Fig. 3. Sensitivities and FPRs of DNN using different percentage of risk factors ranked by contribution scores. Error bar: standard deviation.

The top-ranking 10% risk factors identified by DNN are presented as follows. A02: Age and A03: Gender have highest contributions in successful liver cancer prediction. According to some existing researches, the incidence of liver cancer increases with age [11]. Meanwhile, HCC (main form of liver cancer) is more common among males with a male: female of 2.4 over the world [7]. Therefore, males, especially elderly males should be the focus of liver cancer screening. The contribution score of C01: Air Pollution is high. Little is known about the possible risk associated with exposure to ambient air pollution. A recent research have indicated that exposure to ambient air pollution at residence may increase the risk of liver cancer [30], where nitrogen oxides played a key role. It is noteworthy that if the questionnaire-participants live in big cities, C01: Air Pollution is set true by default. Therefore, in EDTUC, the role of air pollution cannot be clearly concluded. It is clear that people living in big cities have higher risk of developing liver cancer. A07: Occupation indicates that people in different occupations have different possibilities of developing liver cancer (nine occupations considered in EDTUC). However, since we have represent single-choice answers as a scalar, the method we use in this paper only validates the association between liver cancer and occupation in general, but cannot show associations of liver cancer with specific occupation choices. The relationship between specific occupation choices and liver cancer incidence remains to be investigated by further research. Although studies have shown that smoking is a clear risk factor of liver cancer, the rela-tionship between C09: Regular Inhalation of Secondhand Smoke and liver cancer does not seem to be direct. A recent research have found proof that secondhand smoke is an addressable risk factor of nonalcoholic fatty liver disease [31], which is associated with liver cancer. However, that research is mainly for children. Our results suggest that the relationship between secondhand smoke and adult liver cancer should be studied further. Our results confirm that E32: Hyperlipidemia is associated with liver cancer. Hyperlipidemia is a known risk factor for fatty liver [32], which increases HCC incidence significantly in United States [33], and our results suggest that China has similar phenomenon. Epidemiologic evidence strongly suggests that C13: Regular Physical Exercise reduces cancer incidence [16], whereas few studies have focused on relationship between exercise and liver cancer. Our results suggest a strong relationship between liver cancer and exercise. In fact, Regular Physical Exercise reduces a broad range of liver diseases related to liver cancer such as nonalcoholic fatty liver, nonalcoholic steatohepatitis, and Cirrhosis [34], which are known factors that may lead to liver cancer. Our results is in consistent with a recent study who demonstrates that D02: Long-Term Mental Depression has a small positive association with overall occurrence of cancer, and may increase the risk of liver cancer [35]. Psychological factors have been less studied in cancer epidemiology. More studies are required to further research and support these factors. Family history is an important factor in liver cancer (including HCC) occurrence [36]. We find strong associations between F01: Cancer History of Blood Relatives and liver cancer. We notice that F03: Father and F05: Brother are more important than other relatives. We speculate that liver cancer has paternal inheritance characteristic. Using the above nine risk factors, DNN-based LCPM achieves a mean sensitivity of 83.3% and a FPR of lower than 20% when no clinical measures have been introduced.

The top-ranking 10% to 25% risk factors are as follows. E22: Cirrhosis is an intensively-investigated risk factor in liver cancer research. E33: Diabetes have been proved to be a risk factor of liver cancer including HCC, gallbladder and bile duct cancer [37]. E19: Hepatobiliary Disease is an overall risk factor of liver cancer. E09: Other Lung Diseases indicates liver cancer may linked with lung diseases, which are yet to be studied. B04: Coarse Grain, B02: Fresh Fruit, B03: Meat, B09: Pickled Food, and B01: Fresh Vegetable are critical dietary factors associated with liver cancer. These factors appear in the inducing risk factors, indicating that good dietary habits are important to prevent liver cancer. Similar to D02, D01: Recent Mental Trauma is a psychological risk factor that could not be neglected. E01: Hepatitis B Surface Antigen (HBsAg) Test is important in LCPM, which have been confirmed by substantial researches. We also find that E31: Hypertension is associated with liver cancer.

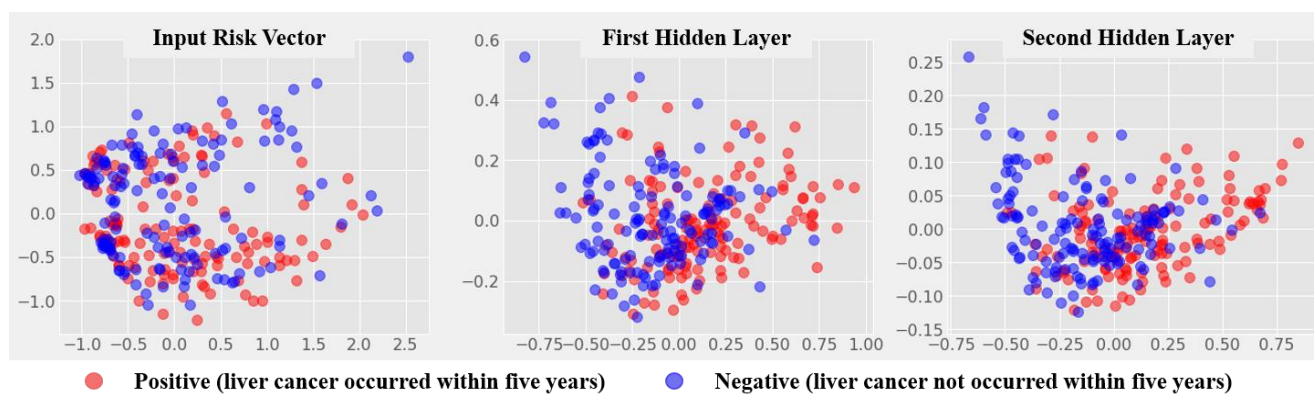

Fig. 4. Representation evolving process in DNN visualized with 2-D PCA technique. From left to right, the representations of positive and negative samples become more linearly separable along with layers going deeper. Each dot stands for a questionnaire sample.

Although most of the identified factors have been supported by existing researches (e.g., gender, age, exercise, cirrhosis, HBsAg, eating habits), there are some less-studied factors (e.g., air pollution, psychological condition, secondhand smoke, occupation, lung diseases). We find these less-studied factors play an important role in decision-making of liver cancer prediction, and thus deserve further study.

The main insufficiency of this work is that we evaluate risk factors of liver cancer only by measuring their contributions in disease prediction, whereas discussions from the perspective of biomedical science are few. We hope this data-driven research could help the continuously exploration of the pathogenesis of liver cancer, so as to facilitate a better prevention system.

For further research, we will continue updating LCPM using up-to-date diagnostic information, and compare results of using statistical analysis method. In addition, we will use machine learning methods to analyze other common cancers (e.g., breast cancer, gastric cancer, colorectal cancer).

## 6 Conclusion

Benefiting from machine learning, LCPM is able to predict liver cancer occurrence within five years. The decision-making of DNN offers insight into the relative contributions of risk factors. Critical risk factors are identified in data-driven and goal-driven approach, which are then validated by liver cancer prediction performance using few top-ranking risk factors. The critical risk factors act as valuable reference to further study and a guideline to healthy lifestyle. This work presents an assistive computing technology for human well-being.

### 7.2 Acknowledgments

This work was supported by Zhejiang Provincial Natural Science Foundation (LQ20F030013). The authors wish to thank Ningbo Health Commission (NHC), China for providing questionnaire data, as well as providing clinical diagnostic records of liver cancer. Corresponding author: Ting Cai.

## References

[1] Torre L A, Bray F, Siegel R L, et al. Global cancer statistics, 2012[J]. CA: a cancer journal for clinicians, 2015, 65(2): 87-108.

[2] Gomaa A I, Khan S A, Toledano M B, et al. Hepatocellular carcinoma: epidemiology, risk factors and pathogenesis[J]. World journal of gastroenterology: WJG, 2008, 14(27): 4300.

[3] Chuang S C, La Vecchia C, Boffetta P. Liver cancer: descriptive epidemiology and risk factors other than HBV and HCV infection[J]. Cancer letters, 2009, 286(1): 9-14.

[4] Larsson S C, Wolk A. Overweight, obesity and risk of liver cancer: a meta-analysis of cohort studies[J]. British journal of cancer, 2007, 97(7): 1005.

[5] Tsukuma H, Tanaka H, Ajiki W, et al. Liver cancer and its prevention[J]. Asian Pacific Journal of Cancer Prevention, 2005, 6(3): 244.

[6] Colditz G A, Atwood K A, Emmons K, et al. Harvard report on cancer prevention volume 4: Harvard Cancer Risk Index[J]. Cancer causes & control, 2000, 11(6): 477-488.

[7] Ghouri Y A, Mian I, Rowe J H. Review of hepatocellular carcinoma: Epidemiology, etiology, and carcinogenesis[J]. Journal of carcinogenesis, 2017, 16.

[8] Ikeda K, Saitoh S, Koida I, et al. A multivariate analysis of risk factors for hepatocellular carcinogenesis: a prospective observation of 795 patients with viral and alcoholic cirrhosis[J]. Hepatology, 1993, 18(1): 47-53.

[9] El-Serag H B, Mason A C. Risk factors for the rising rates of primary liver cancer in the United States[J]. Archives of Internal Medicine, 2000, 160(21): 3227-3230.

[10] Yuen M F, Hou J L, Chutaputti A. Hepatocellular carcinoma in the Asia pacific region[J]. Journal of gastroenterology and hepatology, 2009, 24(3): 346-353.

[11] El‑Serag H B. Epidemiology of hepatocellular carcinoma in USA[J]. Hepatology Research, 2007, 37: S88-S94.

[12] Parkin D M. The global health burden of infection‑associated cancers in the year 2002[J]. International journal of cancer, 2006, 118(12): 3030-3044.

[13] Bowen D G, Walker C M. Adaptive immune responses in acute and chronic hepatitis C virus infection[J]. Nature, 2005, 436(7053): 946.

[14] Hutchinson S J, Bird S M, Goldberg D J. Influence of alcohol on the progression of hepatitis C virus infection: a meta-analysis[J]. Clinical Gastroenterology and Hepatology, 2005, 3(11): 1150-1159.

[15] Kuper H, Tzonou A, Kaklamani E, et al. Tobacco smoking, alcohol consumption and their interaction in the causation of hepatocellular carcinoma[J]. International journal of cancer, 2000, 85(4): 498-502.

[16] Na H K, Oliynyk S. Effects of physical activity on cancer prevention[J]. Annals of the New York Academy of Sciences, 2011, 1229(1): 176-183.

[17] El–Serag H B, Rudolph K L. Hepatocellular carcinoma: epidemiology and molecular carcinogenesis[J]. Gastroenterology, 2007, 132(7): 2557-2576.

[18] LeCun Y, Bengio Y, Hinton G. Deep learning[J]. nature, 2015, 521(7553): 436.

[19] Litjens G, Kooi T, Bejnordi B E, et al. A survey on deep learning in medical image analysis[J]. Medical image analysis, 2017, 42: 60-88.

[20] Shickel B, Tighe P J, Bihorac A, et al. Deep EHR: a survey of recent advances in deep learning techniques for electronic health record (EHR) analysis[J]. IEEE journal of biomedical and health informatics, 2017, 22(5): 1589-1604.

[21] Xiao Y, Wu J, Lin Z, et al. A deep learning-based multi-model ensemble method for cancer prediction[J]. Computer methods and programs in biomedicine, 2018, 153: 1-9.

[22] Kooperberg, Charles, Michael LeBlanc, and Valerie Obenchain. "Risk prediction using genome‑wide association studies." Genetic epidemiology 34.7 (2010): 643-652.

[23] Heidari, Morteza, et al. "Prediction of breast cancer risk using a machine learning approach embedded with a locality preserving projection algorithm." Physics in Medicine & Biology 63.3 (2018): 035020.

[24] Coudray, Nicolas, et al. "Classification and mutation prediction from non–small cell lung cancer histopathology images using deep learning." Nature medicine 24.10 (2018): 1559.

[25] Tyrer, Jonathan, Stephen W. Duffy, and Jack Cuzick. "A breast cancer prediction model incorporating familial and personal risk factors." Statistics in medicine 23.7 (2004): 1111-1130.

[26] Bush, Isabel. "Lung nodule detection and classification." Bush 2016 (2016).

[27] Liao, J. G., and Khew-Voon Chin. "Logistic regression for disease classification using microarray data: model selection in a large p and small n case." Bioinformatics 23.15 (2007): 1945-1951.

[28] Shouman, Mai, Tim Turner, and Rob Stocker. "Using decision tree for diagnosing heart disease patients." Proceedings of the Ninth Australasian Data Mining Conference-Volume 121. Australian Computer Society, Inc., 2011.

[29] Schapire, Robert E. "Explaining adaboost." Empirical inference. Springer, Berlin, Heidelberg, 2013. 37-52.

[30] Pedersen M, Andersen Z J, Stafoggia M, et al. Ambient air pollution and primary liver cancer incidence in four European cohorts within the ESCAPE project[J]. Environmental research, 2017, 154: 226-233.

[31] Lin C, Rountree C B, Methratta S, et al. Secondhand tobacco exposure is associated with nonalcoholic fatty liver disease in children[J]. Environmental research, 2014, 132: 264-268.

[32] Assy N, Kaita K, Mymin D, et al. Fatty infiltration of liver in hyperlipidemic patients[J]. Digestive diseases and sciences, 2000, 45(10): 1929-1934.

[33] Mohamad B, Shah V, Onyshchenko M, et al. Characterization of hepatocellular carcinoma (HCC) in non-alcoholic fatty liver disease (NAFLD) patients without cirrhosis[J]. Hepatology international, 2016, 10(4): 632-639.

[34] Berzigotti A, Saran U, Dufour J F. Physical activity and liver diseases[J]. Hepatology, 2016, 63(3): 1026-1040.

[35] Jia Y, Li F, Liu Y F, et al. Depression and cancer risk: a systematic review and meta-analysis[J]. Public Health, 2017, 149: 138-148.

[36] Turati F, Edefonti V, Talamini R, et al. Family history of liver cancer and hepatocellular carcinoma[J]. Hepatology, 2012, 55(5): 1416-1425.

[37] La Vecchia C, Negri E, Decarli A, et al. Diabetes mellitus and the risk of primary liver cancer[J]. International journal of cancer, 1997, 73(2): 204-207.